%% file: emnlp2023.tex
\pdfoutput=1
\documentclass[11pt]{article}

\usepackage{EMNLP2023}

\usepackage{times}
\usepackage{latexsym}
\usepackage{float}

\usepackage[T1]{fontenc}
\usepackage[utf8]{inputenc}

\usepackage{microtype}

\usepackage{inconsolata}
\usepackage{graphicx}
\usepackage{listings}
\usepackage{dirtytalk}
\usepackage{enumitem}
\usepackage{graphicx}
\usepackage{array}
\usepackage{booktabs}
\usepackage{xcolor}

\usepackage{{booktabs}}
\newcommand{\averitec}{FEVER~8}
\newcommand{\evr}{Ev\textsuperscript{2}R}

\makeatletter
\newcommand\footnoteref[1]{\protected@xdef\@thefnmark{\ref{#1}}\@footnotemark}
\makeatother

\title{AIC CTU@FEVER 8: On-premise fact checking through long context RAG}

\author{Herbert Ullrich \\
AI Center @ CTU FEE\\
Charles Square 13\\
Prague, Czech Republic\\
\texttt{ullriher@fel.cvut.cz} \\\And
Jan Drchal \\
AI Center @ CTU FEE\\
Charles Square 13\\
Prague, Czech Republic\\
\texttt{drchajan@fel.cvut.cz} \\}

\begin{document}
%{\makeatletter\acl@finalcopytrue
  \maketitle
%}
\begin{abstract}
In this paper, we present our fact verification pipeline which has scored first in \averitec{} shared task in real-world automated fact-checking. Our system is a simple two-step RAG pipeline based on our last year's submission. We show how the pipeline can be redeployed on-premise, achieving state-of-the-art fact-checking performance (in sense of \evr{} test-score), even under the constraint of a single Nvidia A10 GPU, 23GB of graphical memory and 60s running time per claim.

\end{abstract}

%%%%%%%%%%%%%%%%%%%%%%%%%%%%%%%%%%%%
% inputs
\input{src/introduction}

\input{src/system_description}
%\input{src/classification}
\input{src/results}
%\input{src/software.tex}
\input{src/conclusions}

%%%%%%%%%%%%%%%%%%%%%%%%%%%%%%%%%%%%

\section*{Limitations}
Our pipeline is not meant to be relied upon nor to replace a human fact-checker, but rather to assist an informed user. It gives sources and proposed labels for further questioning. It is optimized only for English, the carbon costs of the used models are considerable, despite the system trying to cut down the environmental cost of the prediction step.

\section*{Ethics statement}
Our pipeline is an extension of our already existing last year submission all original authors agreed with, including the reusal of the necessary listing in Appendix~\ref{appendix_sec:system_prompt}.
The system was build specifically for the~\averitec~shared task and reflects the biases of its annotators, for more information on this, we suggest the original AVeriTeC paper~\cite{averitec2024}.
\section*{Acknowledgements}
We would like to thank our last year system coauthor Tomáš Mlynář for staying in the loop, providing timely insights, proofreads and experience even when direct participation was not within his time budget this year.

This article was created with the state support of the Ministry of Industry and Trade of the Czech Republic, project no. Z220312000000, within the National Recovery Plan Programme.
The access to the computational infrastructure of the OP VVV funded project CZ.02.1.01/0.0/0.0/16\_019/0000765 ``Research Center for Informatics'' is also gratefully acknowledged.

% Entries for the entire Anthology, followed by custom entries
\bibliography{anthology,custom}
\bibliographystyle{acl_natbib}

\appendix

\include{src/appendix_a_llms}
\include{src/appendix_b_think}

\end{document}

%% file: src/introduction.tex
%!TEX ROOT=../emnlp2023.tex

% show figures/pipeline.png

\section{Introduction}
\label{sec:introduction}
In 2024, Automated Verification of Textual Claims (AVeriTeC) shared task~\cite{schlichtkrull-etal-2024-automated} showed that the fact checking of real-world claims like those from Politifact, AfricaCheck, etc., can be automated to a significant extent, with pipelines accessing Large Language Models (LLMs) to produce the evidence and veracity verdicts for previously unseen claims instead of a human.
Almost each competitive AVeriTeC shared-task system, however, relied on a proprietary LLM like GPT-4o~\cite{rothermel-etal-2024-infact,ullrich-etal-2024-aic} or an open-weights model with high tens of billions of parameters~\cite{yoon-etal-2024-hero}.
This raised a concern -- can the fact-checking process be automated in a way accessible to masses, or is its quality conditioned by the business-owned blackbox models or access to prohibitive computational resources?

In this year's~\averitec{} shared task, the challenge is to match the quality of AVeriTeC systems with ones that only use open-weights models, constrained time of 60 seconds per claim on average, and a fixed compute of a single 23GB A10 GPU.

Our AIC CTU system (Figure~\ref{fig:pipeline}), adapted for \averitec{} from our last year submission, tops its test-leaderboard (Table~\ref{tab:leaderboard}) with a simple Retrieval-augmented Generation (RAG) scheme, using a locally hosted (Ollama) instance of Qwen3 LLM with 14B parameters, leveraging the sheer
%\vspace{-.5em}
\begin{minipage}{\linewidth}
    \centering
    \includegraphics[width=\linewidth]{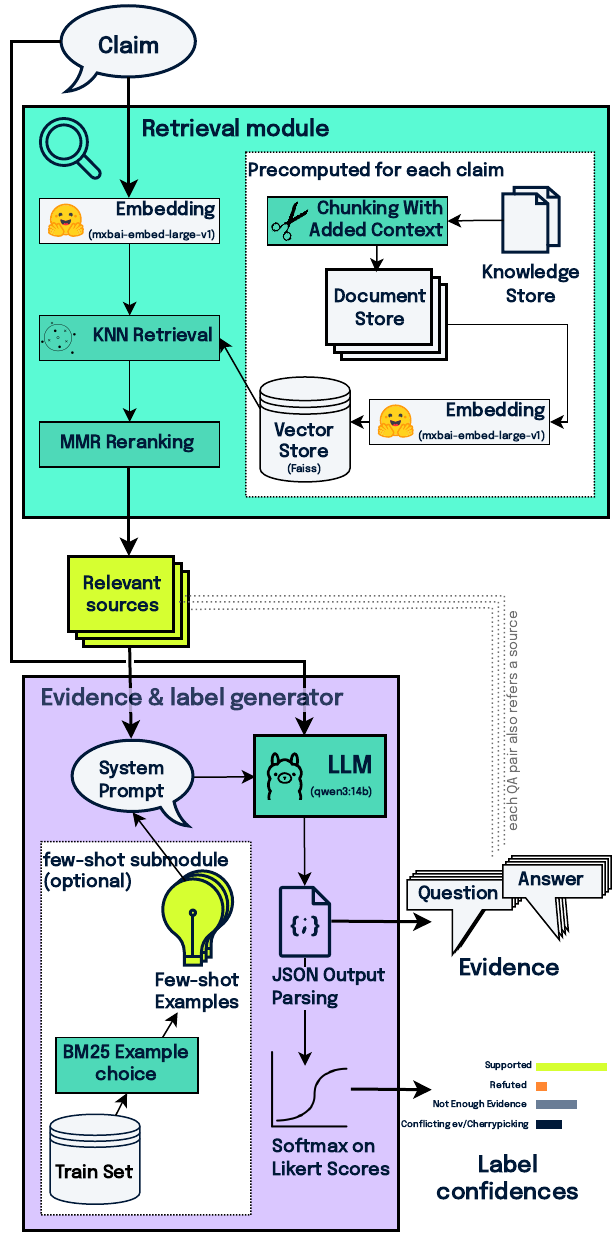}
    \captionof{figure}{Our refreshed fact-checking pipeline used in CTU AIC FEVER 8 submission, adapted from~\citealt{ullrich-etal-2024-aic}.}
    \label{fig:pipeline}
\vspace{1em}
\end{minipage}
context length modern-day LLMs can process.

This paper introduces our system, discusses its design choices and how do they account on the score.
We suggest our system as the new strong baseline -- simple at core, competitive results -- providing the code and reproduction advice.

%\section{Related work}
\label{sec:relwork}
%\begin{enumerate}
%    \item \textbf{\averitec{} shared task}~\cite{averitec2024} releases the datase of real-world fact-checked claims, annotated with evidence available at the date the claim was made.
%\end{enumerate}

%% file: src/system_description.tex
%!TEX ROOT=../emnlp2023.tex

\section{System description}
\label{sec:system}

Our system is a straightforward adaptation of the AIC CTU Averitec system designed one year prior, published in~\citealt{ullrich-etal-2024-aic}.
The cited paper describes the system in detail, with ablation studies and justifications of each step.
Our pipeline, depicted in Figure~\ref{fig:pipeline}, consists of precomputation, retrieval, and generation modules:

\begin{enumerate}[label=\roman*.]  % First-level: i., ii., iii.
\item Precomputation module
\begin{enumerate}[label=\arabic*.]  % Second-level: 1., 2., 3.
    \item The provided AVeriTeC \textbf{knowledge store} \cite{averitec2024} is split into chunks of specified maximum length, each marked with metadata of its URL and the full texts of the chunk before and after.
    \item The chunks are then embedded into their vector representations, using only the chunk texts and no metadata.
    \item Out of all chunk embeddings, a \textbf{vector store} is produced for each claim to be stored as a vector database.
\end{enumerate}
\item Retrieval module
\begin{enumerate}[label=\arabic*.]  % Second-level: 1., 2., 3.
    \item The \textbf{Claim} is embedded into its vector representation using the same model used in i.2.
    \item $k$ nearest neighbours are then retrieved from the vector store, along with their \textbf{chunk embeddings}
    \item The chunk embeddings are then re-ranked using the Maximal Marginal Relevance (MMR) method~\cite{carbonell-mmr}, maximizing the embedding distances between retrieval results while minimizing their distance to the claim.
    Ultimately, we output a subset of $l$~diverse \textbf{sources} for the claim ($l<k$), augmenting each with its context before, after, and the text of its URL.
\end{enumerate}
\item Evidence \& label generation module
\begin{enumerate}[label=\arabic*.]  % Second-level: 1., 2., 3.
    \item We instruct a Large Language Model (LLM) to produce Question-Answer pairs required to fact-check given claim based on the provided sources, and predict its veracity verdict in a single output. We pass it the texts of all $l$ sources, and several few-shot QA-pair generation examples picked from Averitec train set retrieved using BM25 based on the tested claim. The whole instruction is serialized into a system prompt and the format we used can be seen in Appendix~\ref{appendix_sec:system_prompt}.
    \item \textbf{Claim} is then passed to the LLM as a user message.
    \item LLM is called to \textbf{generate the evidence} as a Question-Answer-Source triples and the Likert-scale scores for each possible \textbf{veracity verdict} in a single prediction, performing a chain of thought. 
    \item The LLM output is parsed, and the verdict with the highest score is chosen for the claim.
\end{enumerate}
\end{enumerate}

The main differences between this year's AIC \averitec{} system, opposed to last year's AIC AVeriTeC system, are the omission of knowledge store pruning in the precomputation step\footnote{The precomputed vector stores were required to be independent on claim text in \averitec{}.}, and, importantly, the choice of LLM.
\subsection{Model and parameter choices}
\label{sec:choices}
To produce our submission in the \averitec{} shared task, the following choices were made to deploy the pipeline from section~\ref{sec:system}:

\texttt{mxbai-embed-large-v1}~\cite{li-li-2024-aoe,emb2024mxbai} is used for the vector embeddings, and the maximum chunk size is set to 2048 characters, considering its input size of 512 tokens and a rule-of-thumb coefficient of 4 characters per token to exploit the full embedding input size and produce the smallest possible vector store size without neglecting a significant proportion of knowledge store text.

\texttt{FAISS}~\cite{douze2024faiss,johnson2019billion} index is used as the vector database engine, due to its simplicity of usage, exact search feature and quick retrieval times (sub-second for a single \averitec{} test claim).

$l=10, k=40, \lambda=0.75$ are the parameters we use for the MMR reranking, meaning that 40 chunks are retrieved, 10 sources are yielded after MMR-diversification, and the tradeoff between their similarity to the claim and their diversity is 3:1 in favour of the source similarity to the claim (explained in more detail in~\citealt{ullrich-etal-2024-aic}). 

\texttt{Ollama} wrapper around \texttt{llama.cpp} is the LLM engine we use to deploy LLMs within the \averitec~test environment due to its robustness and ease of deployment.

\texttt{Qwen3-14b}~\cite{yang2025qwen3technicalreport} is the LLM we use to produce the evidence and labels, we also let it generate its own \texttt{<think>} sequences, although further experimentation (Table~\ref{tab:ablation}) suggests that the thinking tokens may not justify the costs of their prediction, as they seem to perform on par with using only the evidence \& label LLM outputs for its chain of thought.

%% file: src/results.tex
%!TEX ROOT=../emnlp2023.tex

\section{Results and analysis}
\label{sec:results}
\label{nothink}

\begin{table}[h]
\centering
\begin{tabular}{l
>{\centering\arraybackslash}p{.7cm} 
>{\centering\arraybackslash}p{.7cm} 
>{\centering\arraybackslash}p{.7cm} 
>{\centering\arraybackslash}p{.7cm} 
>{\centering\arraybackslash}p{.7cm}}
{\small{\textbf{System}}} &
\rotatebox{70}{\textbf{\footnotesize{old AVeriTeC score}}} &
\rotatebox{70}{\textbf{Q only} {\footnotesize{(\evr)}}} &
\rotatebox{70}{\textbf{Q + A} {\footnotesize{(\evr)}}} &
\rotatebox{70}{\textbf{\footnotesize{new AVeriTeC score}}} &
\rotatebox{70}{{\footnotesize{\textbf{time per claim}}}} \\
\hline
{\small{AIC CTU}}       & 0.41 & 0.20 & \textbf{0.48} & \textbf{0.33} & 54\textit{s} \\
{\small{HUMANE}}        & 0.45 & 0.19 & 0.43 & 0.27 & 29\textit{s} \\
{\small{yellow flash}}  & 0.16 & 0.16 & 0.41 & 0.25 & 32\textit{s} \\
{\small{FZIGOT}}        & 0.46 & \textbf{0.36} & 0.40 & 0.24 & 19\textit{s} \\
{\small{EFC}}           & 0.49 & 0.13 & 0.35 & 0.20 & \textbf{~7\textit{s}} \\
{\small{checkmate}}     & 0.38 & 0.18 & 0.34 & 0.20 & 22\textit{s} \\
\hline
{\small{Baseline}}      & \textbf{0.50} & 0.27 & 0.34 & 0.20 & 34\textit{s} \\
\end{tabular}
\caption{\averitec{} shared task system leaderboard as shared by organizers, listing new \evr{}-recall-based~\cite{akhtar2024ev2r} and legacy hu-METEOR AVeriTeC scores. Evaluated using AVeriTeC 2025 test set. Best scores are bold.}
\label{tab:leaderboard}
\end{table}

In Table~\ref{tab:leaderboard}, we reprint the final test-leaderboard of \averitec{} shared task as provided by the organizers.
Our system introduced in Section~\ref{sec:system} scores first in the decisive metric for the task -- the new AVeriTeC score -- with a significant margin.
This came as a surprise to its authors, as neither the values of the old, hu-METEOR-based AVeriTeC score~\cite{averitec2024}, nor the dev-leaderboard available during system development phase (where our system scored 4th), suggested its supremacy.
Let us therefore proceed with a discussion of possible strengths that could have given our system an edge in verifying the \averitec{} test-set of previously unseen 1000 claims.

\subsection{Why does the system perform well?}
\label{sec:why}
So why should our system outperform the \averitec{} baseline and even the other systems submitted to \averitec{} shared task despite the simplicity of its design (Figure~\ref{fig:pipeline}) which boils down to a straightforward case of retrieval-augmented generation (RAG)?

The main reason, in our experience, is the large \textbf{context size} we opt for -- while even the \averitec{} baseline processes the claims and sources in a manner more sophisticated than we do, it processes the knowledge store on a \textit{sentence} level, reducing the amount of information passed to the LLM as opposed to working with \textit{documents} as a whole, which is the strategy our system approximates.

Despite our proposed integration of LLM into the pipeline being rather vanilla, combining sources of total length of as much as 60K characters\footnote{In other words, around 33 standard pages. This number follows from our parameter choices in Section~\ref{sec:choices}: 10 sources are retrieved for each claim, each with $\sim2048$ characters of the embedded text, and additional $\sim4096$ characters of context.} on model input yields highly competitive results, leveraging its own trained mechanisms of context processing.

Our other advantages may have been using a very recent model, Qwen3~\cite{yang2025qwen3technicalreport}, which naturally has a slightly higher leakage of 2025 claims into its train set than older models, and outperforms the previous LLM generations at long sequence processing. Furthermore, our pipeline design only uses a single LLM call per claim, meaning we could use the generously-sized 14B variant of Qwen3 and still match the time limit with Nvidia A10 and 23GB VRAM.

\subsection{Scoring change impact}
\label{sec:score}
While the new AVeriTeC score based on~\evr-recall~\cite{akhtar2024ev2r} estimates the proportion of correctly fact-checked claims\footnote{Claims with sound evidence w.r.t. human annotation, and an exact match in predicted label.} in all claims, just like the old hu-METEOR-based AVeriTeC score did, their underlying methods differ.
Most importantly, an LLM-as-a-judge approach is now used instead of symbolic evidence comparison method.
The rise of our system from 3rd place in AVeriTeC shared task~\cite{schlichtkrull-etal-2024-automated} to 1st place in~\averitec{} without any major system change\footnote{Despite scaling down.} can therefore also be attributed to the used scoring method.
The old scoring method was, for example, found to be prone to some level of noise, as it was not robust against evidence duplication~\cite{malon-2024-multi}, which was a found exploit to boost evidence recall.

The discrepancy between old and new AVeriTeC score in Table~\ref{tab:leaderboard} could motivate a further study on how the new score behaves, for example using the test-prediction files from last year AVeriTeC shared task systems.
The familiarity of the systems, the availability of their hu-METEOR scores and documentation, may reveal valuable insights into the \evr{} evaluation method itself, as in which behaviours does it punish and reward.

\subsection{LLM impact}
\label{llmimp}
\begin{table}[h]
\centering
\begin{tabular}{l
>{\centering\arraybackslash}p{1cm} 
>{\centering\arraybackslash}p{1cm} 
>{\centering\arraybackslash}p{1cm}}
\textbf{LLM} &
\rotatebox{70}{\textbf{Q only} {\footnotesize{(\evr)}}} &
\rotatebox{70}{\textbf{Q + A} {\footnotesize{(\evr)}}} &
\rotatebox{70}{\textbf{\footnotesize{new AVeriTeC score}}} \\
\hline
GPT-4o\textsubscript{\texttt{2024-05-13}}      & 0.30 & 0.58 & 0.40 \\
Llama3.1-70B& 0.37 & 0.54 & 0.39 \\
qwen3:14B\textsubscript{\texttt{/no\_think}}     & 0.29 & 0.59 & 0.41 \\
qwen3:14B\textsubscript{\texttt{/think}}        & 0.20 & 0.59 & 0.42 \\
\hline
\end{tabular}
\caption{Ablation study on LLM choice and \texttt{<think>}-tokens impact on \averitec{} dev-score. Pipeline design (Figure~\ref{fig:pipeline}), retrieval results, system and user prompts are fixed. Evaluated using an on-premise~\evr{} scorer with Ollama-hosted Llama3.3-70B as a judge.}
\label{tab:ablation}
\end{table}

In 2024, we have experimented with then available versions of GPT-4o and Llama3.1-70B and found the open-source Llama to perform encouragingly well, depite the still-quite-cumbersome model size and the need for its quantization~\cite{ullrich-etal-2024-aic}.
This year, we have simply gone for the most recent open-weights LLM at the largest parameter count we could fit within our \averitec{} compute budget, thus choosing the Qwen3 at its 14B parameter size~\cite{yang2025qwen3technicalreport}.

Qwen3 was trained to produce thinking tokens by default, an approach popularized by DeepSeek~\cite{deepseekai2025deepseekr1incentivizingreasoningcapability} and OpenAI research models, to force the chain of thought.
We have experimented with enabling and disabling this feature to see if it has an impact on the AVeriTeC score, and compared the model output quality to our last year prediction dumps, with evaluation experiments listed in Table~\ref{tab:ablation}.

Both Qwen3 evidence and label generation settings perform on par with previous GPT-4o generation, which validates our model choice.
The thinking tokens, while producing legitimate-looking writeups of the fact-checking workflows (see Appendix~\ref{appendix_sec:think}) were not shown to stimulate an improvement in AVeriTeC score in the ablation study (Table~\ref{tab:ablation}), so we suggest to disable this feature in future reproductions in favour of a faster prediction time (54s in the Table~\ref{tab:leaderboard} was produced with the thinking feature \textit{enabled}, so disabling it might solve the issue with near-limit runtime our pipeline suffers from). 

%% file: src/conclusions.tex
%!TEX ROOT=../emnlp2023.tex

\section{Conclusion}
\label{sec:conclusion}
In this paper, we have introduced our simple yet efficient RAG system which performed competitively well under time and compute constraints in \averitec{} shared task, in May 2025.
We release the used code along with usage instructions for producing the~\averitec{} submission, vector stores needed for the pipeline to run and their build scripts at 
\url{https://github.com/heruberuto/FEVER-8-Shared-Task/}
which is a fork of the \averitec{} baseline repository.

We attribute our success mostly to the use of \textit{document} rather than \textit{sentence} level of retrieval granularity and an employment of a recent LLM at a size which utilizes the whole compute and time budget with only around 10\% time reserve as a failsafe.
We encourage further usage of our system as a strong and easy-to-setup baseline for further research in automated fact checking and will be happy to answer any questions on the referred contacts.

\subsection{Future works}
\begin{enumerate}
    \item Integrate a live search API as in~\cite{malon-2024-multi} as a retriever into the AIC pipeline (Figure~\ref{fig:pipeline}) to achieve a real-world generalization
    \item Section~\ref{sec:score} suggests to look at the key differences between legacy and \evr{} scoring methods in terms of the available 2024 AVeriTeC leaderboard and available model documentations -- we believe this could reveal valuable hints both scoring and pipelinne improvements in future work
\end{enumerate}

%% file: src/appendix_a_llms.tex
%!TEX ROOT=../emnlp2023.tex

\lstset{
    language={},
    basicstyle=\ttfamily\footnotesize\linespread{0.9}, % Smaller font with less spacing
    keywordstyle=\color{blue}\bfseries,
    commentstyle=\color{green!50!black}\itshape,
    stringstyle=\color{orange},
    numberstyle=\tiny\color{gray},
    numbers=none, % Line numbers on the left
    stepnumber=1, % Line numbers for every line
    numbersep=5pt, % Space between line numbers and code
    tabsize=4, % Size of tabs
    showstringspaces=false, % Don't show spaces in strings
    breaklines=true, % Line wrapping
    breakatwhitespace=true,
    frame=lines, % Add a frame around the code
    captionpos=b, % Caption at the 
    breakindent=1em,
}
\begin{figure*}
    \section{System prompt}
    \label{appendix_sec:system_prompt}
    \begin{lstlisting}[breaklines=true, language={}, frame=single, caption={System prompt for the LLMs, \averitec{} claim is to be entered into the user prompt. Three dots represent omitted repeating parts of the prompt. Reused and reprinted from~\citealt{ullrich-etal-2024-aic}.}, label={lst:llm_system_prompt}]
You are a professional fact checker, formulate up to 10 questions that cover all the facts needed to validate whether the factual statement (in User message) is true, false, uncertain or a matter of opinion. Each question has one of four answer types: Boolean, Extractive, Abstractive and Unanswerable using the provided sources.
After formulating Your questions and their answers using the provided sources, You evaluate the possible veracity verdicts (Supported claim, Refuted claim, Not enough evidence, or Conflicting evidence/Cherrypicking) given your claim and evidence on a Likert scale (1 - Strongly disagree, 2 - Disagree, 3 - Neutral, 4 - Agree, 5 - Strongly agree). Ultimately, you note the single likeliest veracity verdict according to your best knowledge.
The facts must be coming from these sources, please refer them using assigned IDs:
---
## Source ID: 1 [url]
[context before]
[page content]
[context after]
...

---
## Output formatting
Please, you MUST only print the output in the following output format:
```json
{
 "questions":
     [
         {"question": "<Your first question>", "answer": "<The answer to the Your first question>", "source": "<Single numeric source ID backing the answer for Your first question>", "answer_type":"<The type of first answer>"},
         {"question": "<Your second question>", "answer": "<The answer to the Your second question>", "source": "<Single numeric Source ID backing the answer for Your second question>", "answer_type":"<The type of second answer>"}
     ],
 "claim_veracity": {
     "Supported": "<Likert-scale rating of how much You agree with the 'Supported' veracity classification>",
     "Refuted": "<Likert-scale rating of how much You agree with the 'Refuted' veracity classification>",
     "Not Enough Evidence": "<Likert-scale rating of how much You agree with the 'Not Enough Evidence' veracity classification>",
     "Conflicting Evidence/Cherrypicking": "<Likert-scale rating of how much You agree with the 'Conflicting Evidence/Cherrypicking' veracity classification>"
 },
 "veracity_verdict": "<The suggested veracity classification for the claim>"
}
```
---
## Few-shot learning
You have access to the following few-shot learning examples for questions and answers.:

### Question examples for claim "{example["claim"]}" (verdict {example["gold_label"]})
"question": "{question}", "answer": "{answer}", "answer_type": "{answer_type}"
...
    \end{lstlisting}
\end{figure*}

%% file: src/appendix_b_think.tex
%!TEX ROOT=../emnlp2023.tex

\lstset{
    language={},
    basicstyle=\ttfamily\footnotesize\linespread{0.9}, % Smaller font with less spacing
    keywordstyle=\color{blue}\bfseries,
    commentstyle=\color{green!50!black}\itshape,
    stringstyle=\color{orange},
    numberstyle=\tiny\color{gray},
    numbers=none, % Line numbers on the left
    stepnumber=1, % Line numbers for every line
    numbersep=5pt, % Space between line numbers and code
    tabsize=4, % Size of tabs
    showstringspaces=false, % Don't show spaces in strings
    breaklines=true, % Line wrapping
    breakatwhitespace=true,
    frame=lines, % Add a frame around the code
    captionpos=b, % Caption at the 
    breakindent=1em,
}
\begin{figure*}
    \section{Qwen3 thinking example}
    \label{appendix_sec:think}
    \begin{lstlisting}[breaklines=true, language={}, frame=single, caption={Qwen3 thinking tokens example.}, label={lst:think}]
<think>
Okay, let's tackle this query. The user is claiming that due to Imran Khan's criticism of Macron's comments on Islam, French authorities cancelled the visas of 183 Pakistani citizens and deported 118 from the country. I need to verify this using the provided sources.

First, I'll look through the sources to see if there's any mention of visa cancellations or deportations related to Imran Khan's criticism. 

Looking at Source ID 1, it talks about Imran Khan responding to Macron's remarks and the condemnation of Islamophobic policies, but there's no mention of visa cancellations or deportations. 

Source ID 2 discusses a diplomatic row where Pakistan refused to accept a deportee, Muhammad Mansha, until his credentials were verified. This seems related to a specific case but not the large numbers mentioned in the claim.

Source ID 3 mentions protests in Pakistan against France's republishing of cartoons, but again, no visa or deportation numbers here.

Source ID 4 and 5 discuss diplomatic tensions, but not specific numbers of deportations. Source 5 and 6 from Newscomworld.com talk about fears of deportations of Pakistanis in France but don't mention specific numbers like 183 or 118.

Source ID 7 is about Pakistan deporting French nationals, which is the reverse of the claim. 

Source ID 10 mentions Pakistan summoning the French envoy but again, no numbers on deportations.

None of the sources mention 183 visas cancelled or 118 deported. The closest is Source 2, which mentions one individual, Muhammad Mansha, but that's a single case. The other sources discuss diplomatic tensions, protests, and possible fears of deportations but no actual numbers provided. 

Therefore, the claim about 183 and 118 seems to be unsupported. The answer would be that there's no evidence of such numbers in the sources provided. The veracity verdict would be Refuted, as the sources don't support the specific numbers given.
</think>

```json
{
    "questions": [
        {"question": "Did French authorities cancel visas or deport Pakistani citizens in response to Imran Khan's criticism of Macron's comments on Islam?", "answer": "No specific information about visa cancellations or deportations of 183 Pakistani citizens or 118 deportees is mentioned in the provided sources.", "source": "10", "answer_type": "Boolean"},
        ...
        {"question": "What was the role of social media in the diplomatic dispute between Pakistan and France?", "answer": "Social media was used to spread calls for boycotts of French products and even nuclear attacks on France, but these were not confirmed by official sources.", "source": "5", "answer_type": "Extractive"}
    ],
    "claim_veracity": {
        "Supported": "1",
        "Refuted": "5",
        "Not Enough Evidence": "1",
        "Conflicting Evidence/Cherrypicking": "1"
    },
    "veracity_verdict": "Refuted"
}
```
    \end{lstlisting}
\end{figure*}